\documentclass[article]{elsarticle}

\usepackage{amsmath,amssymb,amsfonts}
\usepackage{graphicx}
\usepackage{textcomp}
\usepackage{multirow}
\usepackage{amsthm}
\usepackage{caption}
\usepackage{subcaption}
\usepackage{layouts}
\usepackage{algorithm}
\usepackage{algorithmic}
\usepackage{tabularx}
\usepackage[export]{adjustbox}
\usepackage{blindtext}
\usepackage[hyphens]{url}
\usepackage{array}
\usepackage{xcolor}
\usepackage{float}
\usepackage{soul}

\newcounter{example}
\newenvironment{example}[1][]{\refstepcounter{example}\par\medskip
   \noindent \textbf{Example~\theexample. #1} \rmfamily}{\medskip}

\newtheorem*{remark}{Remark}

\theoremstyle{definition}
\def\BibTeX{{\rm B\kern-.05em{\sc i\kern-.025em b}\kern-.08em
    T\kern-.1667em\lower.7ex\hbox{E}\kern-.125emX}}

\begin{document}

\begin{frontmatter}

\title{\textsf{DATa}: Domain Adaptation-Aided Deep Table Detection Using Visual--Lexical Representations\tnoteref{mytitlenote}}
\tnotetext[mytitlenote]{This work was supported in part by the National Research Foundation of Korea (NRF) Grant by the Korean Government through MSIT under Grants 2021R1A2C3004345 and 2021R1A4A1029780 and in part by the Institute of Information and Communications Technology Planning and Evaluation (IITP) Grant by the Korean Government through MSIT (6G Post-MAC—POsitioning and Spectrum-Aware intelligenT MAC for Computing and Communication Convergence) under Grant 2021-0-00347.}

\author[kwon,MIDaS]{Hyebin Kwon}
\ead{wdcd6211@yonsei.ac.kr}
\address[kwon]{School of Economics, Yonsei University, Seoul 03722, Republic of Korea}

\author[an,MIDaS]{Joungbin An}
\ead{jbistanbul05@yonsei.ac.kr}
\address[an]{Underwood International College (Nano Science and Engineering), Yonsei University, Seoul 03722, Republic of Korea}
\address[MIDaS]{Machine Intelligence \& Data Science Laboratory, Yonsei University, Seoul 03722, Republic of Korea}

\author[lee]{Dongwoo Lee}
\ead{dongwoolee@skku.edu}
\address[lee]{School of Mechanical Engineering, Sungkyunkwan University, Suwon 16419, Republic of Korea}

\author[shin,po]{{Won-Yong Shin}\corref{mycorrespondingauthor}}
\cortext[mycorrespondingauthor]{Corresponding Author}
\ead{wy.shin@yonsei.ac.kr}
\address[shin]{School of Mathematics and Computing (Computational Science and Engineering), Yonsei University, Seoul 03722, Republic of Korea}
\address[po]{Pohang University of Science and Technology (POSTECH) (Artificial Intelligence), Pohang 37673, Republic of Korea}

\begin{abstract}
Considerable research attention has been paid to table detection by developing not only rule-based approaches reliant on hand-crafted heuristics but also deep learning approaches. Although recent studies successfully perform table detection with enhanced results, they often experience performance degradation when they are used for transferred domains whose table layout features might differ from the source domain in which the underlying model has been trained. To overcome this problem, we present \textsf{DATa}, a novel {\it \underline{D}omain \underline{A}daptation}-aided deep \underline{Ta}ble detection method that guarantees satisfactory performance in a specific target domain where few trusted labels are available. To this end, we newly design {\it lexical features} and an augmented model used for re-training. More specifically, after pre-training \textcolor{black}{one of state-of-the-art vision-based models as our backbone network}, we re-train our augmented model, consisting of \textcolor{black}{the vision-based model} and the multilayer perceptron (MLP) architecture. Using new confidence scores acquired based on the trained MLP architecture as well as an initial prediction of bounding boxes and their confidence scores, we calculate each confidence score more accurately. To validate the superiority of \textsf{DATa}, we perform experimental evaluations by adopting a real-world benchmark dataset in a source domain and another dataset in our target domain consisting of materials science articles. Experimental results demonstrate that the proposed \textsf{DATa} method substantially outperforms competing methods that only utilize visual representations in the target domain. Such gains are possible owing to the capability of eliminating \textcolor{black}{high false positives or false negatives according to the setting of a confidence score threshold}.
\end{abstract}


\end{frontmatter}


\section{Introduction}

\subsection{Background and Motivation}

With the vast majority of information being digitalized, the importance of extracting tables and recognizing their structures from online documents has been growing. For example, materials science is one prominent domain where table detection and structure recognition are critical for comprehensive research. Key information is frequently conveyed in table formats, and extracting such logical and structured information brings forth many benefits.

On the other hand, retrieval of information from the scientific literature provides large-scale training datasets for developing novel materials systems with unprecedented properties using machine learning models~\cite{tshitoyan2019unsupervised}. Materials data in the literature, however, are diverse, unstructured, and heterogeneous, thus making materials data extraction labor-intensive. Although some recently developed text mining models have been proven to successfully acquire materials data from the text, extraction of information from tables in scientific papers is still challenging~\cite{kononova2021opportunities}. With such increasing relevance, the tasks of table detection and structure recognition in digital documents have received significant attention in the literature.

Early studies on the table detection and recognition focused on developing rule-based methods using heuristics based on hand-crafted table layout features~\cite{davulcu2002clustering,long2005model,ng1999learning,jorge2006design} and statistical machine learning (ML) methods such as the support vector machine (SVM) and conditional random field (CRF)~\cite{liu2008identifying}. With the rapid development of deep convolutional neural networks (CNNs), the research paradigm has been shifted towards deep learning-based models. More recent studies have thus focused \textcolor{black}{mostly} on developing deep CNN-based table detection and recognition approaches (see~\cite{agarwal2021cdec,prasad2020cascadetabnet,gilani2017table,huang2019yolo,schreiber2017deepdesrt,paliwal2019tablenet} and references therein), leading to significant performance improvements. By virtue of the availability of a large database with labeled data and automatic feature extraction, almost all state-of-the-art table detection/recognition methods have been in compliance with the development of deep CNNs and their relevant modules.

Nevertheless, table detection still remains a relatively unresolved problem, especially in {\it transferred} domains due to the inconsistency of its form. Even with a steady push to improve the performance using deep learning approaches, table detection poses several practical challenges as follows. First, although scientific documents convey the most essential information in their tabular structure, the form and logic of tables differ immensely among fields, documents, cultures, and even languages, thereby making it harder to build a universal detection model. While state-of-the-art table detection models were trained using well-known large real-world benchmark datasets (e.g., ICDAR 2013~\cite{gobel2013icdar} and ICDAR 2019~\cite{gao2019icdar}) for table detection and recognition, testing the models in a different domain from which they were trained do not often guarantee satisfactory performance due to the inconsistency of the form and logic of tables. Second, unlike well-known benchmark datasets having ground truth tables, lots of documents in specific target domains do not usually contain the ground truth labeling, which is required for model training. These challenges motivate us to design a new methodology of table detection whose performance is not deteriorated even when the underlying model is trained with only a limited amount of data in a target domain with heterogeneity. 

\subsection{Main Contributions}

In this paper, we consider the case where only a small number of documents with ground truth labels are available in a specific target domain of interest, which is a rather feasible scenario. To deal with this case, we introduce \textsf{DATa}, a novel deep table detection method employing the {\it domain adaption} technique so as to guarantee the performance in the target domain whose table layout features might be different from the source domain. In the proposed \textsf{DATa} method, we adopt \textcolor{black}{a} state-of-the-art vision-based model for table detection in the pre-training phase, and build a new augmented model using {\it visual--lexical representations} for re-training.

Before stating our design methodology, we would like to make several empirical findings. First, we show that \textcolor{black}{state-of-the-art methods for {\it general object detection} such as YOLOv5~\cite{glenn_jocher_2020_4154370}, DETR-R50~\cite{detr2020eccv}, and deformable DETR~\cite{deformabledetr2021iclr} outperform state-of-the-art table detection models based on the usage of deep CNNs} in terms of the detection accuracy when popular benchmark datasets are used. Second, more interestingly, \textcolor{black}{assuming a state-of-the-art object detection model is adopted as our backbone network for pre-training using a benchmark dataset in the source domain, we observe that the performance tends to be degraded when the used model is re-trained and tested using another dataset in the transferred domain. Such performance degradation comes from either high {\it false positive} or {\it false negative} rates.} This is most likely due to the fact that not only critical clues from the textual context is ignored in vision-based models but also image-based training renders the underlying model to have a high training data dependency. \textcolor{black}{For example,} the false positives include the cases where graphs, mathematical equations, or even plain text are mistaken as tables. 

\textcolor{black}{In our study, we would like to claim that} many of false positives \textcolor{black}{and false negatives} may have a potential to be eliminated if some {\it lexical} information could be given to the underlying model before final detection. \textcolor{black}{To verify our claim}, we develop our approach that encompasses both vision-based features and lexical features so that it performs successfully in transferred domains having only a few amount of data. More precisely, in \textsf{DATa}, we newly design two types of hand-crafted {\it lexical features}, which include the irregular spacing and the number of table captions. Then, we design an augmented model used for re-training, \textcolor{black}{which consists of an object detection backbone network of choice and the multilayer perceptron (MLP) architecture.} Using new confidence scores acquired based on the trained MLP architecture as well as an initial prediction of bounding boxes and their confidence scores, we show a confidence score calculation algorithm that is capable of calculating each confidence score more accurately.

To validate the superiority of our \textsf{DATa} method, we comprehensively perform empirical evaluations using a real-world benchmark dataset in our source domain and another dataset in a specific target domain. In particular, we collect a number of articles in the field of materials science that serve as the dataset in the target domain. Due to the unavailability of ground truths, we manually annotate the materials science articles so that they are used in the re-training and test phases. Our experimental results demonstrate that the proposed \textsf{DATa} method \textcolor{black}{using object detection models (e.g., YOLOv5, DETR-R50, and deformable DETR)} substantially outperforms \textcolor{black}{its counterpart, i.e., the competing benchmark method that only utilizes visual representations,} in the transferred domain in terms of the detection accuracy. Such gains of \textsf{DATa} are achieved due to the elimination of \textcolor{black}{either false positives or false negatives that tend to occur according to both the setting of a confidence score threshold and the type of object detection models. More specifically, when the confidence score threshold is set low, our \textsf{DATa} method built upon YOLOv5 is beneficial over the pure YOLOv5 model due to the removal of false positives, which is also shown along with visualization of outcomes as case studies. On the other hand, when the confidence score threshold is set high, our \textsf{DATa} method leads to improvements over its counterpart without any lexical features due to the removal of false negatives regardless of the type of object detection models.} Furthermore, we conduct an ablation study to investigate the impact of each lexical feature in the \textsf{DATa} method.

The main technical contributions of this paper are four-fold and summarized as follows:

\begin{itemize}
    \item We propose \textsf{DATa}, a novel table detection method that incorporates a training model for leveraging textual information (i.e., lexical features) into vision-based models.
    \item We show how to design lexical features and an augmented model composed of \textcolor{black}{a general object detection model of choice and the MLP architecture}.
    \item We validate the performance of \textsf{DATa} through extensive experiments using the set of materials science papers. 
    \item We further analyze the impact of lexical features and the parameter sensitivity.
\end{itemize}

\textcolor{black}{To further elaborate on, the advantages of the proposed \textsf{DATa} method are analyzed in three different points of view. First, the superiority of \textsf{DATa} over state-of-the-art models for table detection (including general object detection) comes thanks to the judicious incorporation of lexical features in the underlying backbone network. Second, the performance improvement is achieved with a much smaller training training or pre-training dataset for real-world applications. Unlike the majority of the state-of-the-art methods, \textsf{DATa} is pre-trained only on one small benchmark dataset, named ICDAR 2013, which is one of the smallest, but still exhibits superior performance. Lastly, \textsf{DATa} is highly applicable to transferred domains different from what it was (pre-)trained on. These advantages are especially more significant in the table detection community due to the innate characteristics of tables, where the form and logic tend to change arbitrarily. While humans can intuitively identify tables even without specific captions for them, it is much harder for
machines to do the same with practical challenges.} 

Our \textsf{DATa} method sheds light on how to build a training model in transferred domains having only a small amount of data where no (or few) ground truth label is available so that a broader utilization of table detection is possible in heterogeneous fields.


\subsection{Organization}

The remainder of this paper is organized as follows. In Section 2, we present prior studies related to our work. In Section 3, we explain the methodology of our study, including not only table detection in source and target domains but also an overview of our \textsf{DATa} method. Section 4 describes implementation details of the proposed method. Experimental evaluations are shown in Section 5. Finally, we provide a summary and concluding remarks in Section 6.

\section{Related Work}

Unsurprisingly, with increasing relevance, the tasks of table detection and structure recognition in digital documents have received much attention. However, due to the volatility and variation in style across different document domains, the tasks have remained a struggle with many challenges. So far, a large variety of approaches have been brought forth to aim to perform the two tasks. Since the problem that we tackle in this study is table detection, related studies on table detection are summarized below in two broader research lines, namely, rule-based ML approaches and deep learning approaches.

\textbf{Rule-based ML Approaches.}
Initially, rule-based methods reliant on hand-crafted heuristics and ML algorithms were developed in order to locate tables in the plain text. However, the process of creating many possible features that could indicate a table is not only tedious and inefficient, but also insufficient and domain-dependent, thus making generalization a huge challenge. For example, one of the first approaches to incorporate ML into their models was presented in \cite{ng1999learning} by using specific feature vectors containing spatial and layout information for three different classifiers, each of which is used for detecting the table boundary, columns, and rows. Meanwhile, a clustering-based segmentation routine was presented, and the corresponding table detection model was customized for business letter domains \cite{kieninger2001applying}; the work highlighted difficulties of applicability between domains. Another approach for table detection was designed in \cite{cesarini2002trainable} relying on locating perpendicular lines of white spaces between parallel lines using an MXY tree and then merging located tables. However, such approaches above rendered themselves only applicable to rare domains where tables are always ruled with lines. Later on, a data-driven and flexible method based on the hidden Markov model (HMM) was presented through experiments on combining the transition and node effects of the Markov chain for table location. An SVM was also employed to identify column and row separators along with 26-dimensional hand-crafted low-level features. 

\textbf{Deep learning approaches.}
In realizing that a more elaborate and sophisticated method is necessary, coupled with the rapid development of deep CNNs, more recent studies have focused on using deep CNNs to successfully perform table detection with enhanced results. 

Hao et al.\cite{hao2016table} was one of the first to incorporate CNNs that are run over some region proposals selected using a loose set of rules. Into their model; some non-visual information (metadata) such as rendering instructions and character properties was also leveraged to assist detection. Afterwards, faster R-CNN and image transformation techniques were quickly brought into the field of table detection \cite{gilani2017table}. A multi-scale multi-task fully convolutional network (FCN) was presented in \cite{he2017multi} for segmenting tables, where individual tables were extracted using some heuristic rules and a verification network to reduce false positives.
Later on, it was recognized that models could benefit from deploying pre-trained weights from other large-scale deep CNN models that were actively applied for image classification and object detection in natural scenery images. Thus, reaping the benefits of transfer learning, researchers have begun to initialize their models with these pre-trained weights and have made use of low-level feature detectors. It was shown in \cite{schreiber2017deepdesrt} how to fine-tune a pre-trained faster R-CNN model. The advantages of transfer learning were also leveraged in \cite{siddiqui2018decnt} by using a pre-trained feature pyramid network (FPN) for faster R-CNN and adapting deformable convolutional layers in order to account for geometric transformation and a variation of scales.
More recent attempts to detect tables have utilized the modules from the field of instance segmentation that not only locate and classify objects but also classify images down to the pixel-level. A cascade mask R-CNN model was initiated in \cite{agarwal2021cdec} by incorporating a composite (dual) backbone. A more sophisticated cascade mask R-CNN model was developed in \cite{prasad2020cascadetabnet} along with a high resolution based backbone (HRNet); iterative transfer learning and customized image augmentation techniques were shown to further enhance the performance. \textcolor{black}{Additionally, recent studies include variations of cascade R-CNN; in~\cite{fernandes2022tabledet}, cascade R-CNN with a complete Intersection-Over-Union (IOU) loss and a deformable convolution backbone was presented to capture the variations in scales and orientations of tables, and the cascade R-CNN with a
deformable convolution backbone was also leveraged in~\cite{nguyen2022cdersnet} to solve object
detection in Vietnamese documents with a Rank \& Sort (RS) loss. In~\cite{nguyen2022tablesegnet}, an FCN with different paths was designed to detect tables from
a high or low resolution image. Moreover, new datasets such as UIT-DODV
(the first Vietnamese image document dataset) and TNCR~\cite{abdallah2022tncr} (a dataset with varying image quality collected from free open source websites) have been added to the benchmark datasets of the table detection and segmentation community.}

\textcolor{black}{Meanwhile, the use of transformers has also gained attention in the computer vision community. A direct application of transformers to image recognition was shown in~\cite{vit2020iclr}. Detection transformer (DETR)~\cite{detr2020eccv} provided end-to-end object detection with remarkable results. Recent powerful object detection and segmentation
models built upon DETR such as deformable DETR~\cite{deformabledetr2021iclr} and conditional
DETR~\cite{conditionaldetr2021iccv} were also presented in order to address the slow convergence issue of DETR, where various techniques including deformable attention
and a better design of queries were developed.}

\textbf{Discussion.}
Despite recent achievements based on \textcolor{black}{deep learning models for table detection}, a common problem that \textcolor{black}{most of} studies struggle to effectively evade is getting rid of \emph{false positives}, which are errors incorrectly indicating the presence of a table when the table does not indeed exist and are inevitably quite frequent due to low inter-class variability. More precisely, this implies that there are other classes of document elements (such as figures, graphs, images, charts, etc.) that have similar layout features, which are thus not readily differentiated from tables visually. In prior studies, this obstacle was overcome by increasing training iterations and using larger datasets. 
Even with enhancing the capability of generalization and applicability, models still tend not to perform appropriately when switching up to a specific target domain (e.g., the materials science) that often lacks in the training data. This motivates us to come up with a novel design methodology along with \textit{domain adaptation} for efficiently reducing false positives whilst utilizing transfer learning to enhance the performance in a specific target domain from a benchmark source domain. 

\section{Methodology}

In this section, we describe how tables are detected not only in source domains using benchmark datasets but also in a specific target domain whose table layout features might be different from the source domains. Next, we explain an overview of the proposed \textsf{DATa} method using visual--lexical representations.

\subsection{Table Detection in Source Domains} \label{SEC:Benchmark}

Lots of research efforts have been made in developing state-of-the-art table detection models based on the usage of deep \textcolor{black}{learning approaches} by virtue of the availability of benchmark datasets such as the ICDAR 2013, ICDAR 2019, and Marmot datasets. In this subsection, we revisit the performance comparison among existing table detection methods using such benchmark datasets. The datasets contain document files from different sources, where the detailed description of these datasets is given below. 

{\bf ICDAR Table Competition 2013} \cite{gobel2013icdar}. The dataset consists of 238 pages in the PDF format, corresponding to European Union and US Government reports.

{\bf ICDAR Table Competition 2019} \cite{gao2019icdar}. This dataset contains not only modern documents but also hand-written table images and historical documents.

{\bf Marmot} \cite{agarwal2021cdec}. This dataset is a collection of 2,000 pages in the PDF format, which is composed of Chinese and English documents at the proportion of about 1:1. While the Chinese pages were selected from over 120 e-Books, the English pages were collected from the Citeseer website.

Before showing the performance comparison among state-of-the-art methods of table detection for several benchmark datasets, let us state three metrics, which are widely used to evaluate the accuracy of table detection. The \textit{precision} is defined as the ratio of relevant predictions over all predictions and shows how many tables predicted by a given model are actually real ones. The \textit{recall} is defined as the ratio of relevant predictions over all instances and measures how many tables a given model finds over all the tables present in the dataset. The $F_1$ score is the harmonic mean of the precision and recall. 

\begin{table}[t!]
\centering
\scriptsize 
\caption{Performance comparison among state-of-the-art methods \textcolor{black}{for table detection} in terms of three performance metrics when three benchmark datasets are used.}
\begin{tabular}{ |c|c|c|c|c| }
\hline
Dataset & Model & Precision & Recall & $F_1$ Score \\
\hline\hline
\multirow{6}{5em}{ICDAR 2013} & YOLOv5 \cite{glenn_jocher_2020_4154370} & 1.0 & 1.0 & \textbf{1.0} \\
& CascadeTabNet \cite{prasad2020cascadetabnet} & 1.0 & 1.0 & \textbf{1.0} \\
& DeCNT \cite{siddiqui2018decnt} & 0.996 & 0.996 & 0.996 \\
& DeepDeSRT \cite{schreiber2017deepdesrt} & 0.9615 & 0.9740 & 0.9677 \\
& TableNet \cite{paliwal2019tablenet} & 0.9628 & 0.9697 & 0.9662 \\
& CDeC-Net \cite{agarwal2021cdec} & 0.942 & 0.993 & 0.968 \\
& \textcolor{black}{DETR-R50} \cite{detr2020eccv} & \textcolor{black}{0.967} & \textcolor{black}{0.971} & \textcolor{black}{0.969} \\ 
& \textcolor{black}{Deformable DETR} \cite{deformabledetr2021iclr} & \textcolor{black}{0.966} & \textcolor{black}{0.971} & \textcolor{black}{0.968} \\ 
\hline
\multirow{3}{5em}{ICDAR 2019} & YOLOv5 \cite{glenn_jocher_2020_4154370} & 0.980 & 0.975 & \textbf{0.978} \\
& CDec-Net \cite{agarwal2021cdec} & 0.930 & 0.971 & 0.950 \\
& TableRadar \cite{gao2019icdar} & 0.940 & 0.950 & 0.945 \\
& \textcolor{black}{DETR-R50} \cite{detr2020eccv} & \textcolor{black}{0.946} & \textcolor{black}{0.971} & \textcolor{black}{0.958} \\
& \textcolor{black}{Deformable DETR} \cite{deformabledetr2021iclr} & \textcolor{black}{0.969} & \textcolor{black}{0.994} & \textcolor{black}{0.981} \\ 
\hline
\multirow{3}{5em}{Marmot} & YOLOv5 \cite{glenn_jocher_2020_4154370} & 0.955 & 0.940 & \textbf{0.947} \\
& DeCNT \cite{siddiqui2018decnt} & 0.946 & 0.849 & 0.895 \\
& CDeC-Net \cite{agarwal2021cdec} & 0.779 & 0.943 & 0.861 \\
& \textcolor{black}{DETR-R50} \cite{detr2020eccv} & \textcolor{black}{0.914} & \textcolor{black}{0.945} & \textcolor{black}{0.929} \\
& \textcolor{black}{Deformable DETR} \cite{deformabledetr2021iclr} & \textcolor{black}{0.932} & \textcolor{black}{0.991} & \textcolor{black}{0.931} \\ 
\hline
\end{tabular}
\label{table:1}

\end{table}

Now, we comprehensively summarize the performance of table detection for well-known available benchmark datasets. For each dataset, we split it into two subsets: 80\% as the training set and 20\% as the test set. In Table 1, we present the performance of state-of-the-art methods \textcolor{black}{(including general object detection models)} for table detection with respect to three performance metrics using three real-world benchmark datasets, including ICDAR 2013, ICDAR 2019, and Marmot. The parameters for each method were tuned in such a way that the best performance is achieved. In the table, the value highlighted in bold indicates the best performer for each \textcolor{black}{dataset with respect to the $F_1$ score}. From Table 1, our findings are as follows:

\begin{itemize}
    \item All table detection models exhibit quite satisfactory performance.
    \item The YOLOv5 model is \textcolor{black}{generally the best performer regardless of the dataset}. 
    \item \textcolor{black}{Other general object detection models such as DETR-R50 and deformable DETR perform quite satisfactorily while achieving more than 0.9 of the $F_1$ score for all datasets.}
\end{itemize}
\textcolor{black}{From these findings, we adopt three object detection models (i.e., YOLOv5, DETR-R50, and deformable DETR) as our backbone network,} which will be pre-trained using a benchmark dataset.


\subsection{Table Detection in Target Domains}

Due to the fact that key information in document files is frequently conveyed in table formats, extracting such logical and structured information brings forth significant benefits. However, building and training a new table detection model suit for a target domain would be ineffective, costly, and non-contributory. This is because very few (or no) trusted labels are available in a target domain of interest. Hence, it is a practical challenge how to design a table detection model that should work appropriately for a target domain using \textit{domain adaptation} techniques. In our study, we focus on the materials science as our target domain since it is one of prominent areas where table mining plays a crucial role in carrying out comprehensive research along with summaries of key properties of materials extracted from tables.

We would like to investigate the fundamental limit of the performance on the table detection in target domains after training an underlying model from different sources. To this end, we evaluate the performance of \textcolor{black}{general object detection models} trained with benchmark datasets for table detection when labeled materials science articles are used as a test set (whose description will be shown in Section 5.1). Despite the fact that our test set in a target domain (i.e., the field of materials science) is a relatively small group of document pages containing tables that are easily identifiable and follow existing formats similarly, our experiments reveal high false positives \textcolor{black}{and false negatives} while going through document files. \textcolor{black}{For example,} the predictions are often incorrect especially when there are either random blocks of parallel text chunks shaped in rectangular form or even mathematical equations separate from the plain text. Some examples of the false positives are displayed in Figure 1 and are elaborated on in the following.

\begin{figure}[t!]
    \centering
    \begin{subfigure}[]{0.65\textwidth}
        \centering
        \includegraphics[width = \linewidth]{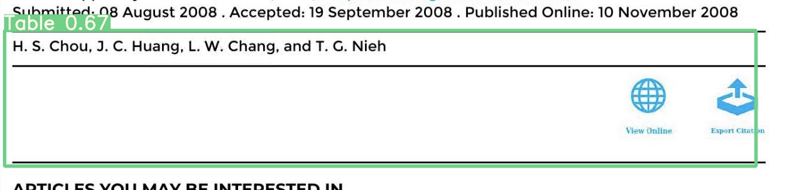}
        \caption{The case where a part of the cover was misclassified as a table with the confidence score of 0.57.}
        \label{fig: my first subfigure}
    \end{subfigure}
    \break
    \break
    \begin{subfigure}[]{0.65\textwidth}
        \centering
        \includegraphics[width = \linewidth]{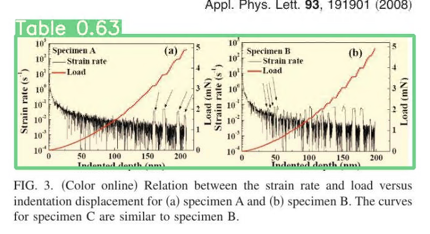}
        \caption{The case where a figure was misclassified as a table with the confidence score of 0.63.}
        \label{fig: my second false positive}
    \end{subfigure}
    \break
    \break
    \begin{subfigure}[]{0.65\textwidth}
        \centering
        \includegraphics[width = \linewidth]{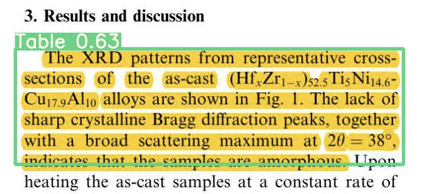}
        \caption{The case where a highlighted text area was misclassified as a table with the confidence score 0.63.}
        \label{fig: my third false positive}
    \end{subfigure}
    \caption{Three cases of false positives in our target domain.}
    \label{fig: 1}
\end{figure}

\begin{example}
Figure 1 shows three cases of false positives in our target domain (i.e., the field of materials science) \textcolor{black}{when the YOLOv5 model is used for table detection}, where each subfigure includes a bounding box, meaning the rectangular box that binds the object of interest (i.e., a table), and a confidence score, meaning a probability between 0 and 1 that indicates how likely each bounding box is to be a table.
In Figure 1-(a), it is seen that a part of the cover in an article was misclassified as a table. This comes from the fact that the part includes parallel horizontal ruling lines and irregularities in the text. In Figure 1-(b), we observe that a figure was misclassified as a table due to similar reasons to those in Figure 1-(a). In Figure 1-(c), even normal, the plain text with the yellow highlighting tool is shown to be misclassified as a table because of the horizontal lines and the partial clusters that the highlighting tool created. While all three examples have significantly high confidence scores of the table existence as depicted in Figure 1, these results imply that vision-based features extracted from state-of-the-art deep \textcolor{black}{learning} models do not precisely capture all distinguishable patterns in the target domain. 
\end{example}

\textcolor{black}{Examples of false negatives in our target domain can also be investigated although they are not shown in this paper.}
We note that an inherent characteristic of tables in documents is their low inter-class and high intra-class variability. This means that, while it is easy to misclassify tables as other document class elements, the form of the table itself is highly variable. Although the \textcolor{black}{general object detection model such as YOLOv5} has a sophisticated architecture with multiple layers, simply training the model using benchmark datasets available for table detection would not be effective. In other words, it is apparent that there are limitations when we only utilize vision-based approaches for transferred domains having a limited number of labeled datasets due to the intra-class variability. This motivates us to design a new deep table detection method via domain adaption using not only vision features but also additional features, namely lexical features.

\subsection{Our \textsf{DATa} Method}

In this subsection, we explain our methodology along with the pipeline of the proposed \textsf{DATa} method. The overall procedure, composed of the \textit{pre-training} and \textit{re-training} phases, is described as follows.

\begin{itemize}
    \item \textbf{Step 1.} We train \textcolor{black}{an object detection model as a backbone network} using a prevalent benchmark dataset, which corresponds to the pre-training phase. \textcolor{black}{Note that the chosen backbone network for our model must produce the standard output format:} an initial prediction of bounding box coordinates and their confidence scores.
    \item \textbf{Step 2.} The \textcolor{black}{chosen backbone network} is re-trained using another dataset in a target domain (e.g., a set of materials science articles) in order to extract vision-based features.
    \item \textbf{Step 3.} To extract (hand-crafted) \textit{lexical} features, we crop the original PDF file based on \textcolor{black}{the backbone network's} predicted bounding box coordinates that we obtained in Step 2. 
    \item \textbf{Step 4.} We build a separate MLP architecture that is to be trained with feature vectors built upon lexical features. The MLP returns its own confidence score.
    \item \textbf{Step 5.} We then acquire a new confidence score based on the trained \textcolor{black}{backbone network} and the MLP architecture.
    \item \textbf{Step 6.} In the test phase, we run our method including the two trained models to predict bounding box coordinates and their confidence scores.
\end{itemize}

\begin{figure*}[t!]
    \includegraphics[width=1\linewidth]{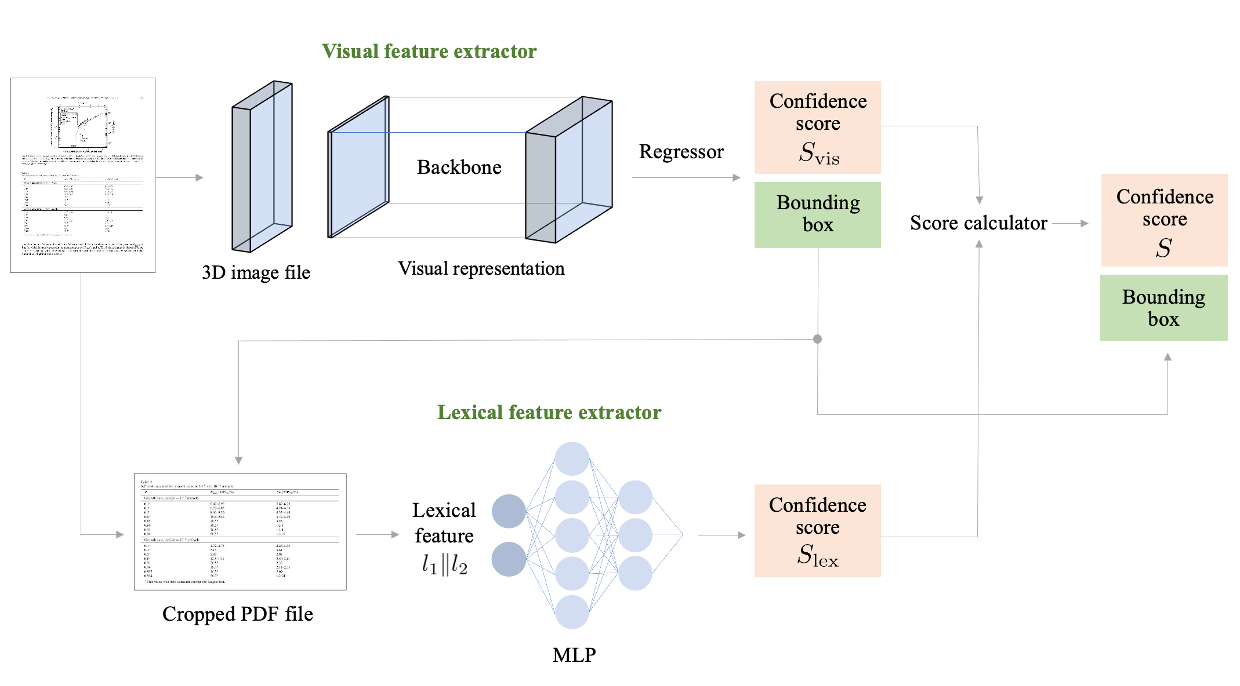}
    \caption{The schematic overview of our \textsf{DATa} method.}
    \label{fig:2}
\end{figure*}

The schematic overview of our \textsf{DATa} method is illustrated in Figure 2. As depicted in the figure, the input format of the \textcolor{black}{object detection backbone network} and the MLP architecture is given by image and PDF files, respectively. In the MLP architecture, we additionally add lexical features extracted from the cropped PDF file as input. Since we aim at predicting confidence scores more accurately by leveraging lexical representations, we attempt to make a final prediction from the two confidence scores of both \textcolor{black}{the chosen backbone network} and the MLP architecture in the context of domain adaptation. The implementation details will be shown in the next section. The benefits of our \textsf{DATa} method are summarized below. 

\begin{remark}
In the re-training phase, an additional module (i.e., the MLP architecture) is augmented as a separate network trained only on lexical representations. Thus, such a module is light-weight and convenient to train. Furthermore, the proposed method is \textcolor{black}{deep-learning-model-agnostic} due to the fact that our backbone network can be replaced with other competing deep \textcolor{black}{learning} models \textcolor{black}{as long as an initial prediction of bounding box coordinates and their confidence scores can be produced as the standard output format}.
\end{remark}

\section{Implementation Details of the \textsf{DATa} Method}

In this section, we elaborate on the design of lexical features and an augmented training model for table detection. Recall that \textcolor{black}{a backbone network of choice} is initially pre-trained on a benchmark dataset and then re-trained with a dataset from our target domain (i.e., the materials science), while predicting confidence scores and bounding box coordinates, as depicted in Figure 2. We focus primarily on both how to create lexical features used as input of the MLP architecture and how to design an augmented model based on the MLP for re-training.

\subsection{Design of Lexical Features}

Although deep \textcolor{black}{object detection} models satisfactorily perform table detection, they still pose some practical challenges especially when we run the models in transferred domains. The design of lexical features is motivated by the observation that our own experimental results from training \textcolor{black}{such object detection} models only with vision-based features tend to exhibit \textcolor{black}{either high false positives or high false negatives}, which will be empirically shown in Section 5. This is inevitable due to the low inter-class and high intra-class variability of tables in document files. To overcome this problem, we present hand-crafted lexical features to be added to the backbone network, which are inspired by prior studies on rule-based ML approaches for table detection (see \cite{pinto2003table,peng2006information,zanibbi2004survey,ng1999learning,liu2008identifying} and references therein).

To this end, we determine which features are the most prominent and effective in the sense of distinguishing tables from non-tables by investigating document files not only in the source domain but also in our target domain (i.e., the Material Science). In consequence, we take into account two types of lexical features that bring the synergy effect with each other in reinforcing the table detection performance. Specifically, the first feature is \textit{irregular spacing} $l_1$, which indicates the number of lines such that irregular spacings are contained in bounding boxes of interest. The second feature is the number of \textit{table captions}, denoted as $l_2$. Then, the resulting feature to be fed into the MLP architecture is given by $l_1 \| l_2$, where $\|$ is the vector concatenation operator. 

\subsubsection{Irregular Spacing}

To design the lexical feature $l_1$, we start by making our empirical findings that represent distinguishable patterns of tables in document files. Specifically, we make the following insightful observations from document files:

\begin{enumerate}
\item Tables tend to have at least one line whose number of (rowwise) irregular spacings is more than two.
\item Tables tend to have a couple of consecutive lines, each of which contains more than three irregular spacings.
\end{enumerate}

Based on the above observations, we would like to establish the following two criteria by using two thresholds $n_\text{space}>0$ and $n_\text{line;1}>0$:

\begin{itemize}
\item If the number of irregular spacings in one row is greater than $n_\text{space}$, then we treat the row as the line that is {\it potentially} within a table, termed the \textit{relevant line}. 
\item  Check whether there are irregularly spaced relevant lines, whose pattern is similar to each other, in the range of $n_\text{line;1}$ lines to account for consecutiveness.
\end{itemize}
The feature $l_1$ is finally given by the total number of consecutive irregularly spaced relative lines in the given cropped PDF file that fulfill the above two criteria.

Note that if $n_\text{space}$ is set too low, then the possibility that the final count of irregular spaces includes false positives increases. For example, column margins could be falsely counted for double or triple spaced documents, which is not the type of irregular spacings that we want to find. Conversely, if the threshold is set too high, then the final count could be missing some false negatives. 

Additionally, while random irregular spacings often occur in a line for several reasons (e.g., the case where the corresponding line is a part of a mathematical equation or a figure), regular irregular spacings that continue for a couple of consecutive lines can be a distinguishable feature that tables usually have. In this context, we need to carefully determine the values of $n_\text{space}$ and $n_\text{line;1}$, which will be specified in Section 5.

\subsubsection{Table Caption}

To design another lexical feature $l_2$, from academic articles such as materials science articles, we start by making our empirical finding that all tables are likely to accompany their caption somewhere nearby. Thus, adding a ``table caption''-related feature would be quite advantageous in improving the detection accuracy.

While the feature $l_1$ indicates the total number of consecutive irregularly spaced relevant lines in the given PDF file, we are now interested in the \textit{indices} (i.e., positions) of such consecutive irregularly spaced relevant lines in the same PDF file. We use a threshold $n_\text{line;2}>0$ to identify the existence of a table caption, where the value of $n_\text{line;2}>0$ will be specified in Section 5. Then, we search for not only the consecutive irregularly spaced relevant lines but also $n_\text{line;2}>0$ more lines before and after them to make sure whether there is a table caption. Note that if $n_\text{line;2}$ is set too high, then the function could start counting false positives, where the word ``table'' is used in another context rather than as a caption (although this is carefully avoided in our setting). The feature $l_2$ is finally given by the number of table captions in the given PDF file.

\subsection{Design of the Augmented Model for Re-training}

In this subsection, we describe our augmented model used for re-training, which consists of two networks including the \textcolor{black}{object detection backbone network of choice} and the MLP architecture, whose input formats are given by image and PDF files, respectively. We also show a confidence score calculation algorithm, described in Algorithm 1, that chooses the final confidence score for table detection in the test phase. 

\subsubsection{The MLP Architecture}

In addition to the confidence score, denoted as $S_\text{vis}$, achieved by vision-based features, we are interested in calculating another confidence score relying on lexical features designed in Section 4.1. To this end, we use the MLP architecture as follows:

\begin{align}
S_\text{lex} = MLP(l_1 \| l_2),
\end{align}
where $S_\text{lex}$ is the confidence score based on the two-dimensional lexical feature vector $l_1 \| l_2$.

\subsubsection{The Confidence Score Calculation Algorithm}

\begin{algorithm}[t]
\caption{: Confidence score calculation}
\begin{algorithmic}
 \renewcommand{\algorithmicrequire}{\textbf{Input:}}
 \renewcommand{\algorithmicensure}{\textbf{Output:}}
 \REQUIRE $\theta > 0$
 \ENSURE  $S_\text{final}$
  \IF {$S_\text{tex} \geq \theta$}
  \IF {$S_\textsf{tex} \ge S_\text{vis}$}
  \STATE {$conf \leftarrow S_\text{lex}$}
  \ENDIF
  \ELSIF {$S_\text{lex} < \theta$}
  \STATE {$conf \leftarrow S_\text{vis}$}
  \ENDIF
  \RETURN $S_\text{final}$
\end{algorithmic}
\end{algorithm}

Due to the fact that two confidence scores $S_\text{vis}$ and $S_\text{lex}$ are available, we need to take advantage of them for more accurately detecting tables. To this end, we present a confidence score calculation algorithm such that our augmented model chooses between the two confidence scores with a certain threshold $\theta>0$.

If the confidence score $S_\text{lex}$ is higher than both the pre-defined $\theta$ and the original $S_\text{vis}$, then our final confidence score $S$ is chosen as $S_\text{lex}$, which would potentially result in more accurate detection. Otherwise, our final confidence score $S$ is set as $S_\text{vis}$. The threshold $\theta$ is appropriately chosen via empirical validations, which shall be shown in Section 5.

\section{Experimental Evaluation}

In this section, we first describe real-world datasets used in the evaluation. Then, we describe performance metrics and benchmark methods. After describing our experimental settings, we comprehensively evaluate the performance of our \textsf{DATa} method and the most competing benchmark method.

\subsection{Datasets}

To pre-train the backbone network (i.e., \textcolor{black}{YOLOv5~\cite{glenn_jocher_2020_4154370}, DETR-R50~\cite{detr2020eccv}, and deformable DETR~\cite{deformabledetr2021iclr}}) in the source domain, we adopt ICDAR 2013 out of three real-world benchmark datasets addressed in Section 3.1. This is because ICDAR 2013 is the most widely used dataset in the field of table detection and recognition where every state-of-the-art model mentioned above has universally adopted this dataset. Moreover, the ICDAR 2013 dataset provides PDF documents in its initial format before image conversion---the PDF file is the input format of the MLP architecture in our \textsf{DATa} method. 

In addition to the benchmark dataset in the source domain, we adopt another dataset in our target domain, which is a collection of articles in the field of materials science. The dataset includes only parts of the research articles published from 1994 to 2019 in the metallurgy field while describing the composition and microstructure-dependent properties of alloys. The dataset contains 279 pages in the PDF format with varying layouts. The corresponding ground truths were custom-labeled. That is, we manually annotated a set of materials science articles for table detection, which is used in the study as an example of a transferred domain from a broader dataset. Retrieval of information from the scientific literature provides large-scale materials data for machine learning models to design novel materials with enhanced properties~\cite{tshitoyan2019unsupervised, kononova2021opportunities}. We use the dataset in the re-training and test phases.

\subsection{Performance Metrics and Benchmark Methods}

As already stated in Section~\ref{SEC:Benchmark}, to validate the performance of the \textsf{DATa} and state-of-the-art methods, we adopt the precision, recall, and $F_1$ score as performance metrics of table detection.

Now, let us state state-of-the-art methods for performance comparison with our \textsf{DATa} method. As addressed in Section 3.1, there are lots of state-of-the-art methods for table detection, which include but not limited to YOLOv5~\cite{glenn_jocher_2020_4154370}, \textcolor{black}{
DETR-R50 \cite{detr2020eccv}, 
deformable DETR \cite{deformabledetr2021iclr},} 
CascadeTabNet~\cite{prasad2020cascadetabnet},  DeCNT~\cite{siddiqui2018decnt}, DeepDeSRT~\cite{schreiber2017deepdesrt}, TableNet~\cite{paliwal2019tablenet}, CDeC-Net~\cite{agarwal2021cdec}, and TableRadar~\cite{gao2019icdar}. As summarized in Table 1, \textcolor{black}{general object detection models (i.e., YOLOv5, DETR-R50, and deformable DETR) are shown to outperform other state-of-the-art table detection methods} for all benchmark datasets (i.e., ICDAR 2013, ICDAR 2019, and Marmot). \textcolor{black}{Thus, in our \textsf{DATa} method, we use the three object detection models as our backbone network.}


\subsection{Experimental Setup}

\textcolor{black}{From Table 1, we recall that the YOLOv5 model is generally the best performer. Thus, we consider the YOLOv5x model (the largest and heaviest one out of four YOLOv5 models including YOLOv5s, YOLOv5m, YOLOv5l, YOLOv5x) for pre-training unless otherwise stated.}
 Specifically, YOLOv5 is trained with 1,200 epochs, with a batch size of 4 after reshaping each input image file to a size of $640\times640\times3$. All other parameters and functions in YOLOv5 are used in default. 
 
 In the pre-training phase, we split each benchmark dataset in a source domain into training/validation sets with a ratio of 80/20\%. In the re-training phase, the custom-labeled materials science articles are used; the \textcolor{black}{used object detection} model \textcolor{black}{as our backbone network} is re-trained with 200 epochs with a batch size of 1, and all other parameters and functions are left identical to the pre-training settings. We split the dataset in our target domain (i.e., the materials science) into (re-)training/test sets with a ratio of 80/20\%.

Next, we describe how to set parameters used for extracting lexical features that are fed into our augmented model based on the MLP architecture. As default settings, three thresholds $n_\text{space}$, $n_\text{line;1}$, and $n_\text{line;2}$ are given by 3, 2, and 7, respectively, unless otherwise specified. We shall empirically analyze the parameter sensitivity in the next subsection.

For the confidence score calculation algorithm, the value of the threshold $\theta$ is empirically found. Our augmented model chooses one between two confidence scores $S_\text{vis}$ and $S_\text{lex}$ alongside the given $\theta$. This enables us to correct some imperfections of the augmented model implemented by the MLP architecture. In other words, the parameter $\theta$ is used to ensure that erroneous predictions are not taken into account. We empirically found that $\theta=0.3$ is almost optimal in the sense of maximizing the $F_1$ score, which is thus used for our experiments. 

Furthermore, we define another threshold $\eta>0$ \textcolor{black}{, the so-called {\it confidence score threshold},} in order to finally detect a table in comparison with the predicted confidence score $S$. The case of $S\ge\eta$ is regarded as the existence of a table; on the other hand, the case of $S<\eta$ is treated as the table non-existence. In our study, we tune the parameter $\eta$ differently for each experiment to better explain the corresponding experimental results, unless otherwise stated.

\subsection{Experimental Results}

In this subsection, our empirical study is designed to answer the following four research questions. 

\begin{itemize}
\item {\it RQ1.} How much does the \textsf{DATa} method improve the performance on table detection over the benchmark method without any lexical features?

\item{\it RQ2.} How does the \textsf{DATa} method provide an interpretation of its benefit?

\item {\it RQ3.} How much does each lexical feature in the \textsf{DATa} method contribute to the performance?

\item {\it RQ4.} How do model parameters affect the performance of the \textsf{DATa} method?
\end{itemize}

To answer these questions, we comprehensively carry out experiments in the following.

\subsubsection{Comparison with the Benchmark Method (RQ1)}

The performance comparison between our \textsf{DATa} methods \textcolor{black}{using YOLOv5, DETR-R50, and deformable DETR as a backbone network (corresponding to \textsf{DATa+YOLO}, \textsf{DATa+DETR}, and \textsf{DATa+deDETR}, respectively)} and \textcolor{black}{the three competing object detection models} as state-of-the-art methods for table detection is presented in Table 2 with respect to performance metrics such as the precision, recall, and $F_1$ score using the datasets in source and target domains. 
The performance is evaluated according to different values of the threshold $\eta$ ranging from 0.4 to 0.9, which is used for finally detecting a table in comparison with $S$. 
In Table 2, the value highlighted in bold indicates the best performer for each case \textcolor{black}{with respect to the $F_1$ score}. We would like to make the following observations:

\begin{table}[t!]
    \centering
    \scriptsize 
    \caption{Performance comparison between the proposed \textsf{DATa} method and \textcolor{black}{the three state-of-the-art methods for table detection} in terms of the precision, recall, and $F_1$ score according to different values of threshold $\eta$ in finally detecting a table.}
    \begin{tabular}{|c|c|c|c|c|c|}
         \hline
         Data split (\%) & $\eta$ & Method & Precision & Recall & $F_1$ Score \\
         \hline\hline
        \multirow{10}{3em}{80:20} & \multirow{2}{2em}{0.4} & YOLOv5 & 0.875 & 1.0 & 0.934 \\
        & & \textcolor{black}{DETR-R50} & \textcolor{black}{1.0} & \textcolor{black}{0.769} & \textcolor{black}{0.870} \\
        & & \textcolor{black}{Deformable DETR} & \textcolor{black}{1.0} & \textcolor{black}{0.785} & \textcolor{black}{0.880} \\
        & & \textcolor{black}{\textsf{DATa+YOLO}} & 1.0 & 1.0 & \textbf{1.0} \\ 
        & & \textcolor{black}{\textsf{DATa+DETR}} & \textcolor{black}{1.0} & \textcolor{black}{0.769} & \textcolor{black}{0.870} \\
        & & \textcolor{black}{\textsf{DATa+deDETR}} & \textcolor{black}{1.0} & \textcolor{black}{0.929} & \textcolor{black}{0.963} \\
        \cline{2-6} 
        & \multirow{2}{2em}{0.5} & YOLOv5 & 0.975 & 1.0 & 0.987 \\
        & & \textcolor{black}{DETR-R50} & \textcolor{black}{1.0} & \textcolor{black}{0.769} & \textcolor{black}{0.870} \\ 
        & & \textcolor{black}{Deformable DETR} & \textcolor{black}{1.0} & \textcolor{black}{0.785} & \textcolor{black}{0.880} \\
        & & \textcolor{black}{\textsf{DATa+YOLO}} & 1.0 & 1.0 & \textbf{1.0} \\
        & & \textcolor{black}{\textsf{DATa+DETR}} & \textcolor{black}{1.0} & \textcolor{black}{0.769} & \textcolor{black}{0.870} \\
        & & \textcolor{black}{\textsf{DATa+deDETR}} & \textcolor{black}{1.0} & \textcolor{black}{0.929} & \textcolor{black}{0.963} \\
        \cline{2-6}
        & \multirow{2}{2em}{0.6} & YOLOv5 & 0.934 & 1.0 & 0.965 \\
        & & \textcolor{black}{DETR-R50} & \textcolor{black}{1.0} & \textcolor{black}{0.769} & \textcolor{black}{0.870} \\ 
        & & \textcolor{black}{Deformable DETR} & \textcolor{black}{1.0} & \textcolor{black}{0.785} & \textcolor{black}{0.880} \\
        & & \textcolor{black}{\textsf{DATa+YOLO}} & 1.0 & 1.0 & \textbf{1.0} \\
        & & \textcolor{black}{\textsf{DATa+DETR}} & \textcolor{black}{1.0} & \textcolor{black}{0.769} & \textcolor{black}{0.870} \\
        & & \textcolor{black}{\textsf{DATa+deDETR}} & \textcolor{black}{1.0} & \textcolor{black}{0.929} & \textcolor{black}{0.963} \\
        \cline{2-6}
        & \multirow{2}{2em}{0.7} & YOLOv5 & 0.934 & 1.0 & 0.965 \\
        & & \textcolor{black}{DETR-R50} & \textcolor{black}{1.0} & \textcolor{black}{0.769} & \textcolor{black}{0.870} \\ 
        & & \textcolor{black}{Deformable DETR} & \textcolor{black}{1.0} & \textcolor{black}{0.714} & \textcolor{black}{0.833} \\
        & & \textcolor{black}{\textsf{DATa+YOLO}} & 1.0 & 1.0 & \textbf{1.0} \\
        & & \textcolor{black}{\textsf{DATa+DETR}} & \textcolor{black}{1.0} & \textcolor{black}{0.769} & \textcolor{black}{0.870} \\
        & & \textcolor{black}{\textsf{DATa+deDETR}} & \textcolor{black}{1.0} & \textcolor{black}{0.929} & \textcolor{black}{0.963} \\
        \cline{2-6}
        & \multirow{2}{2em}{0.8} & YOLOv5 & 1.0 & 0.857 & 0.923 \\
        & & \textcolor{black}{DETR-R50} & \textcolor{black}{1.0} & \textcolor{black}{0.769} & \textcolor{black}{0.870} \\ 
        & & \textcolor{black}{Deformable DETR} & \textcolor{black}{1.0} & \textcolor{black}{0.714} & \textcolor{black}{0.833} \\
        & & \textcolor{black}{\textsf{DATa+YOLO}}& 1.0 & 0.857 & 0.923\\
        & & \textcolor{black}{\textsf{DATa+DETR}} & \textcolor{black}{1.0} & \textcolor{black}{0.769} & \textcolor{black}{0.870} \\
        & & \textcolor{black}{\textsf{DATa+deDETR}} & \textcolor{black}{1.0} & \textcolor{black}{0.929} & \textcolor{black}{\bf 0.963} \\
        \cline{2-6}
        & \multirow{2}{2em}{0.9} & YOLOv5 & 1.0 & 0.571 & 0.727 \\
        & & \textcolor{black}{DETR-R50} & \textcolor{black}{1.0} & \textcolor{black}{0.629} & \textcolor{black}{0.818} \\ 
        & & \textcolor{black}{Deformable DETR} & \textcolor{black}{1.0} & \textcolor{black}{0.714} & \textcolor{black}{0.833} \\
        & & \textcolor{black}{\textsf{DATa+YOLO}} & 1.0 & 0.857 & 0.923 \\
        & & \textcolor{black}{\textsf{DATa+DETR}} & \textcolor{black}{1.0} & \textcolor{black}{0.769} & \textcolor{black}{0.870} \\
        & & \textcolor{black}{\textsf{DATa+deDETR}} & \textcolor{black}{1.0} & \textcolor{black}{0.929} & \textcolor{black}{\bf 0.963} \\
        \hline
    \end{tabular}
\end{table}

\begin{itemize}
\item Our \textsf{DATa} method consistently and substantially outperforms the state-of-the-art methods regardless of threshold values. Such a gain is possible since the incorporation of lexical features is able to complement missing parts of information from what was retrieved only using vision-based representations.

\item \textcolor{black}{The mechanism behind the proposed \textsf{DATa} method can be traced down to the fact that the confidence score can be more accurately calculated in both ways. In other words, confidence scores are lowered when \textsf{DATa} detects no table, and confidence scores are calculated near 1.0 when \textsf{DATa} does detect a table. Thus, not only are the false positives reduced, but the false negatives are reduced as well, leading to an overall performance improvement.}

\item \textcolor{black}{When the confidence score threshold $\eta$ is set low (e.g., $\eta=0.4$), \textsf{DATa+YOLO} is beneficial over YOLOv5. In this case, the YOLOv5 model experiences performance degradation in terms of the precision. This is due to the existence of some {\it false positives} when YOLOv5 is tested in the target domain composed of materials science articles.}



\item \textcolor{black}{On the other hand, when $\eta$ is set high (e.g., $\eta=0.9$), our \textsf{DATa} method leads to improvements over its counterpart that only uses visual representations. In this case, all the benchmark methods tend to get degraded in terms of the recall. This degradation comes from the existence of high {\it false negatives} when the benchmark methods are tested in the target domain.}


\item The performance gap between our \textsf{DATa} method ($X$) and \textcolor{black}{its counterpart without any lexical feature ($Y$) is the largest in terms of the $F_1$ score when YOLOv5 is used as a backbone network and $\eta=0.9$; the maximum improvement rate of $26.96\%$ is achieved}, where the improvement rate (\%) is given by $\frac{X-Y}{Y}\times100$.


\item \textcolor{black}{The advantages of using \textsf{DATa+DETR} are less clear than other results. This is due to the fact that, while our \textsf{DATa} method adjusts the confidence score, it relies on the original bounding box predictions completely. Thus, if the initial bounding box prediction is not precise, then the impact and benefits of \textsf{DATa} can be reduced. In our experiments, this is the case where DETR-R50 is used as a backbone network.}

\item As long as the parameter $\eta$ is properly set (e.g., $\eta\in[0.4,0.7]$), \textcolor{black}{the \textsf{DATa+YOLO} method achieves the perfect detection scores in terms of the precision and recall while not exhibiting any false positives and false negatives}.
\end{itemize}

\textcolor{black}{In subsequent experiments, we assume our \textsf{DATa} method with YOLOv5 as its backbone network. That is, the \textsf{DATa} method is regarded as \textsf{DATa+YOLO} from now on.}

\subsubsection{Interpretation with Case Studies (RQ2)}

To more closely interpret how the \textsf{DATa} method is beneficial, we conduct case studies alongside visualization of outcomes. 
Two examples visualizing the output for both the benchmark and \textsf{DATa} methods \textcolor{black}{(i.e., YOLOv5 and \textsf{DATa+YOLO})} are shown in Figure 3, where the bounding box and the resulting confidence score are depicted for each case when a certain image/PDF file in the test set is used as input.

\begin{figure}[t!]
    \centering
    \begin{subfigure}[h]{0.65\textwidth}
        \centering
        \centering\includegraphics[width=9cm]{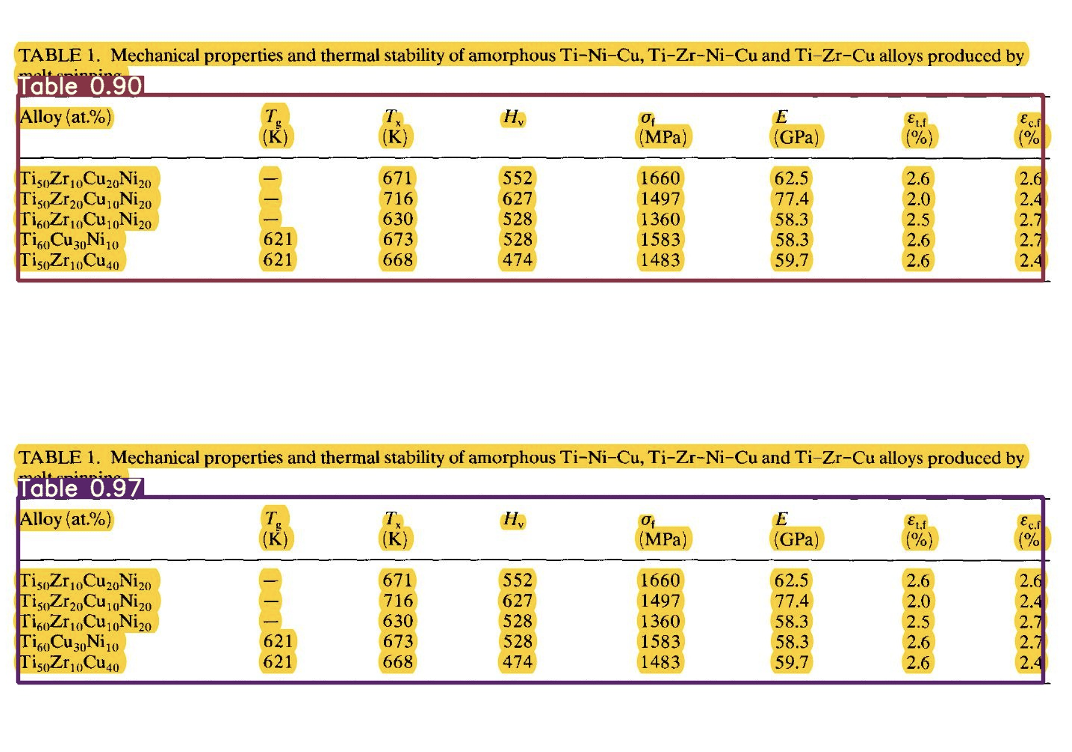}
        \caption{{\bf Case 1} where the confidence score of a true positive is further improved using the \textsf{DATa} method.}  
        \label{fig : mspaper1_page4 subfigure}
    \end{subfigure}
    \centering
    \break
    \break
    \centering
    \begin{subfigure}[h]{0.65\textwidth}
        \centering
        \includegraphics[width=8.5cm]{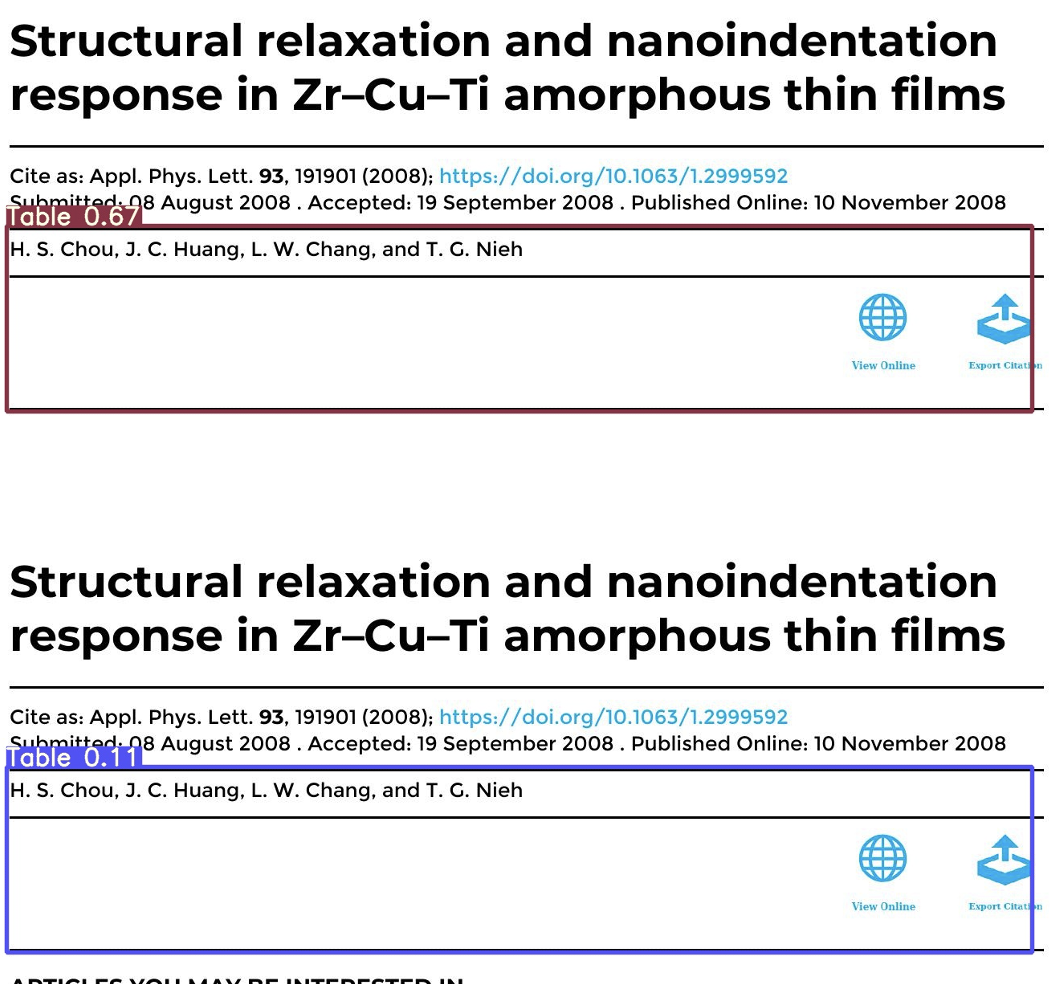}
        \caption{{\bf Case 2} where a false positive is captured and its confidence score is significantly reduced using the \textsf{DATa} method.}
        \label{fig : mspaper8_page1 subfigure}
    \end{subfigure}
    \centering
    \label{fig : another figure}
    \caption{Examples visualizing the output for both \textcolor{black}{the pure YOLOv5 model and the \textsf{DATa+YOLO} method} as a form of the bounding box and the resulting confidence score.}
\end{figure}

\begin{example}
In Figure 3-(a), the top and bottom images show the bounding box and the predicted confidence score from \textcolor{black}{the pure YOLOv5 model and the \textsf{DATa+YOLO} method}, respectively. It is seen that the confidence score of a true positive is further improved by using the \textsf{DATa} method.
\end{example}

\begin{example}
As another example, in Figure 3-(b), the top and bottom images show the bounding box and the predicted confidence score from \textcolor{black}{the pure YOLOv5 model and the \textsf{DATa+YOLO} method}, respectively. The outcome of the YOLOv5 model yields a false positive where a non-tabular component is mistaken as a table with a confidence score of 0.67. However, from the outcome of the \textsf{DATa} method, the false positive is no longer recognized as a table with a considerably reduced confidence score.
\end{example}

\subsubsection{Impact of Lexical Features (Ablation Studies) (RQ3)}

Next, in order to discover what role each lexical feature plays in the success of the \textsf{DATa} method, we conduct an ablation study by removing each feature in our method \textcolor{black}{(i.e., \textsf{DATa+YOLO})}.

\begin{itemize}
\item \textsf{DATa}: This corresponds to the original \textsf{DATa} method without removing any lexical features.

\item \textsf{DATa-$l_1$}: The lexical feature $l_1$ related to irregular spacings is removed. Then, the confidence score $S_\text{lex}$ is calculated by the MLP with the feature $l_2$ as input, i.e., $S_\text{lex}=MLP(l_2)$.

\item \textsf{DATa-$l_2$}: The lexical feature $l_2$ related to table captions is removed. Then, the confidence score $S_\text{lex}$ is calculated as $S_\text{lex}=MLP(l_1)$.
\end{itemize}

The performance comparison among the original \textsf{DATa} and its two variants \textcolor{black}{removing one of lexical features} is presented in Figure 4 with respect to the $F_1$ score. From the figure, our findings are as follows.

\begin{itemize}
\item In Figure 4-(a), when $\eta=0.9$, the original \textsf{DATa} method exhibits potential gains over other variants, which demonstrate that each lexical feature plays a crucial role together in detecting tables.

\item More interestingly, when $\eta=0.9$, the performance gap between \textsf{DATa} and \textsf{DATa-$l_1$} is much lower than that between \textsf{DATa} and \textsf{DATa-$l_2$}. In other words, \textsf{DATa-$l_1$} is superior to \textsf{DATa-$l_2$}. This finding indicates that the feature $l_2$ related to table captions is more valuable than that of irregular spacings. Such a tendency come from the fact that the feature $l_2$ implicitly contains the information related to irregular spacings to some extent since the design of $l_2$ leverages the outcome from the irregular spacing discovery.

\item In Figure 4-(b), for $\eta=0.7$ (corresponding to the case where the perfect detection score is achieved by the original \textsf{DATa}), the performance gap between \textsf{DATa} and its two variants gets reduced significantly compared to the case of $\eta=0.9$. This implies that removing a lexical feature in the \textsf{DATa} method leads to no or little performance degradation.
\end{itemize}

\begin{figure}[t!]
    \centering
    \begin{subfigure}[h]{0.65\textwidth}
        \centering
        \includegraphics[width=\linewidth]{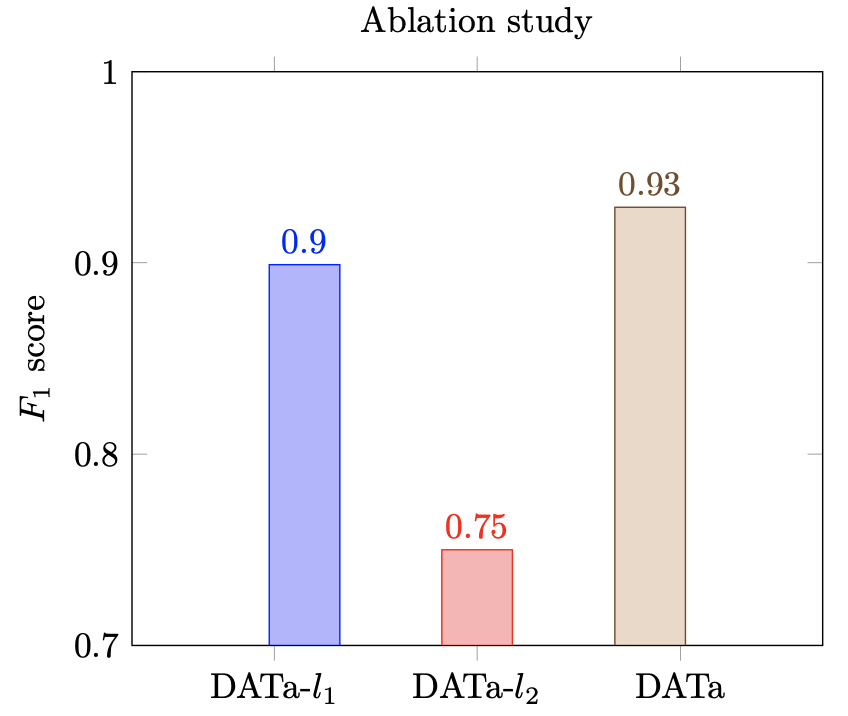}
        \caption{$\eta=0.9$}
    \end{subfigure}
    \break
    \break
    \centering
    \begin{subfigure}[h]{0.65\textwidth}
        \centering
        \includegraphics[width=\linewidth]{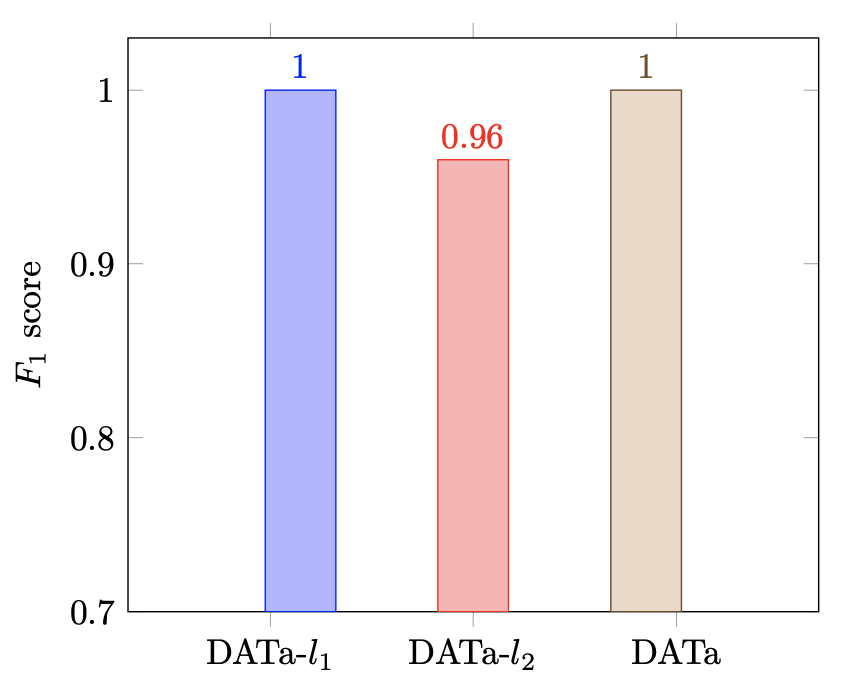}
        \caption{$\eta=0.7$}
    \end{subfigure}
    \caption{Performance comparison among the \textsf{DATa} method and its two variants in terms of the $F_1$ score.}
\end{figure}

\subsubsection{Sensitivity Analysis (RQ4)}

\begin{figure}[t!]
    \centering
    \begin{subfigure}[h]{0.51\textwidth}
        \centering
        \includegraphics[width=\linewidth]{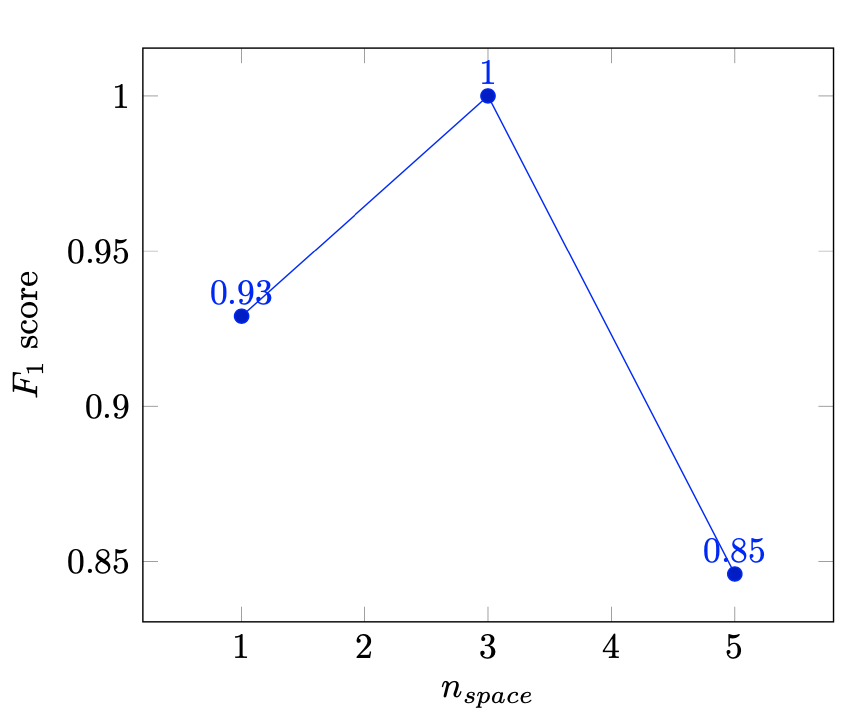}
        \caption{$F_1$ score versus $n_\text{space}$.}  
        \label{fig5}
    \end{subfigure}
    \break
    \break
    \begin{subfigure}[h]{0.51\textwidth}
        \centering
        \includegraphics[width=\linewidth]{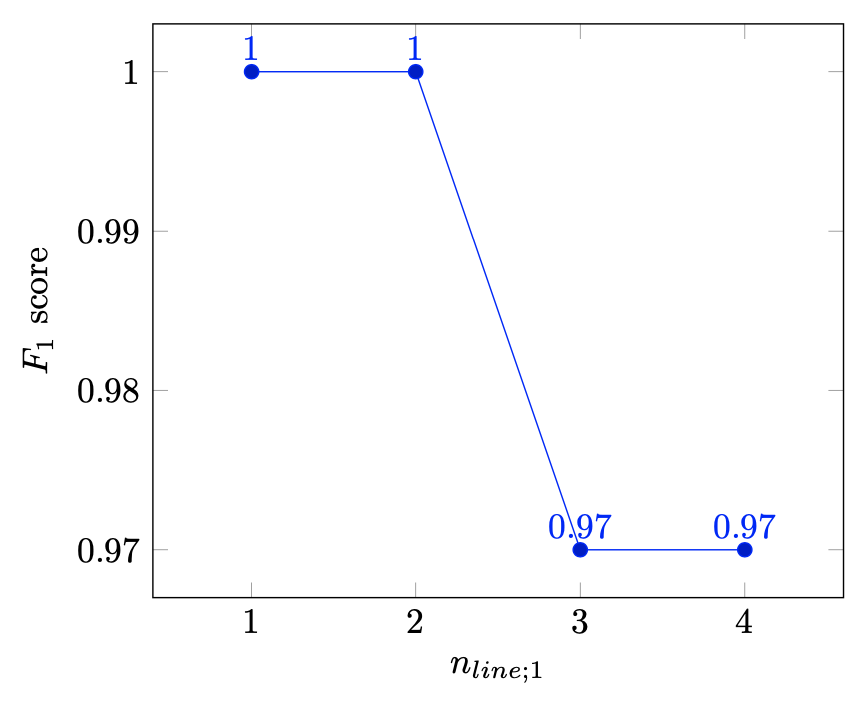}
        \caption{$F_1$ score versus $n_\text{line;1}$.}
     \label{fig6}
    \end{subfigure}
    \break
    \break
    \begin{subfigure}[h]{0.51\textwidth}
        \centering
        \includegraphics[width=\linewidth]{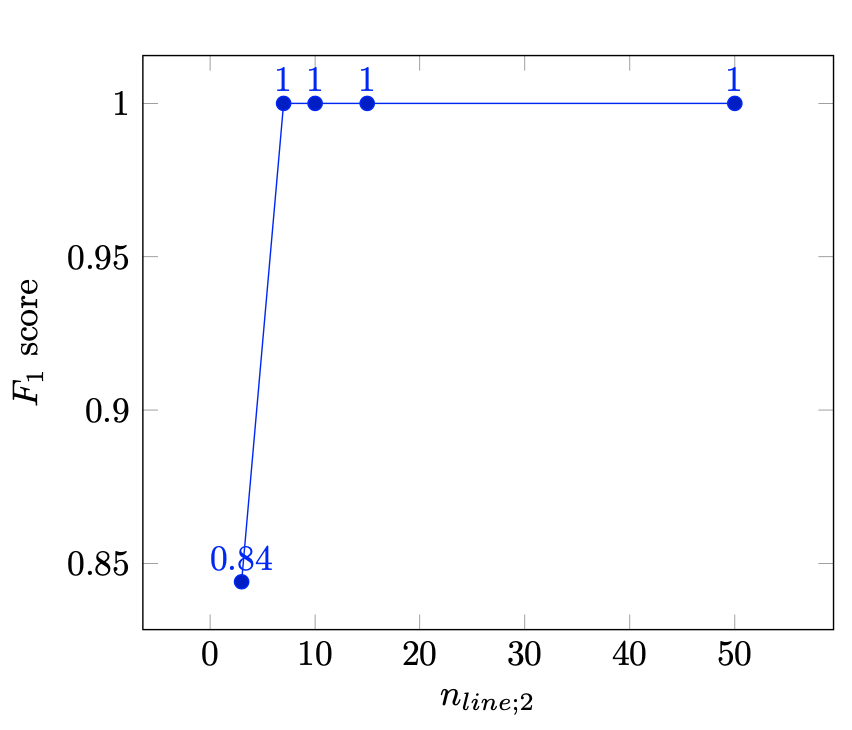}
        \caption{$F_1$ score versus $n_\text{line;2}$.}
     \label{fig7}
    \end{subfigure}
    \label{fig5-7}
    \caption{$F_1$ score according to different values of each threshold in designing lexical features.}
\end{figure}

In Figure~5, we investigate the impact of three parameters $n_\text{space}$, $n_\text{line;1}$ and $n_\text{line;2}$ on the performance of \textcolor{black}{\textsf{DATa+YOLO}} in terms of the $F_1$ score, where $\eta=0.3$ is used. Here, $n_\text{space}$ refers to the threshold to determine whether each line is the relevant line, having more than $n_\text{space}$ irregular spacings; $n_\text{line;1}$ is the range for which we check irregular consecutiveness; and $n_\text{line;2}$ is the number of lines before and after consecutive irregularly spaced relevant lines to make sure whether there is a table caption. When a parameter ($n_\text{space}$, $n_\text{line;1}$, or $n_\text{line;2}$) varies so that its effect is clearly revealed, other parameters are set to their pivot values $n_\text{space}=3$, $n_\text{line;1}=2$, and $n_\text{line;2}=7$. We make the following observations:

\begin{itemize}
\item From Figure 5-(a), we observe that the maximum $F_1$ score is achieved at $n_\text{space}=3$. Thus, it is important to set the value of $n_\text{space}$ appropriately. 
\item Figure 5-(b) shows that the best value for $n_\text{line;1}$ is either 1 or 2 and the $F_1$ score gets deteriorated as the value of $n_\text{line;1}$ further grows. This is because, as the range for which we check consecutive irregularities increases, the probability that the lines are indeed consecutive decreases accordingly. Furthermore, as our default setting, the value of 2 was chosen over 1 since lines that are truly consecutive in a table will be often read not to be digitally, due to other unexpected elements on the page.
\item From Figure 5-(c), it is seen that the performance consistently approaches the maximum $F_1$ score beyond a certain value of $n_\text{line;2}$.\footnote{This finding may not be always exhibited, but other behaviors may be observed depending on test datasets. This is because the performance would often be deteriorated for large $n_\text{line;2}$ due to the fact that table captions that do not belong to the appointed region could be falsely counted.} However, it may not be desirable to set $n_\text{line;2}$ to a large value due to the computational complexity.
\end{itemize}

\section{Concluding Remarks}
This paper introduced a novel table detection method, termed \textsf{DATa}, that guarantees satisfactory performance in transferred domains where very few trusted labels are available. In the \textsf{DATa} method, a domain adaptation technique was employed using visual--lexical representations. \textcolor{black}{Motivated by the fact that 1) there are likely to be high false positive and false negative rates when purely vision-based table detection models are used in different domains and 2) there is often the lack of training data in transferred domains, we newly designed hand-crafted lexical features.} We also developed an augmented model used for re-training, \textcolor{black}{consisting of two networks including a state-of-the-art vision-based model as our backbone model and MLP architecture}. Through comprehensive experiments using a real-world benchmark dataset in the source domain and another dataset in the target domain (i.e., the materials science), we demonstrated that 1) the proposed \textsf{DATa} method substantially outperforms \textcolor{black}{competing benchmark methods, including YOLOv5, DETR-R50, and deformable DETR,} that only use visual representations in our specific target domain, 2) the maximum improvement rate of \textcolor{black}{26.96\% over YOLOv5 is achieved in terms of the $F_1$ score when \textsf{DATa-YOLO} as the best performer is used}, and 3) \textcolor{black}{the \textsf{DATa-YOLO} method indeed achieves the perfect detection scores} in terms of the precision and recall as long as the \textcolor{black}{confidence score} threshold $\eta$ is set appropriately. Moreover, we investigated how much each lexical feature in the \textsf{DATa} method contributes to the performance along with an ablation study. 
    
\textcolor{black}{\textsf{DATa} is advantageous compared to other table detection methods in three different perspectives. First, the superiority of \textsf{DATa} over state-of-the-art vision-based models comes due to the fact that the lexical representations are capable of capturing information that cannot be extracted by purely vision-based models; this information plays a crucial role in reducing either false positives or false negatives according to the setting of parameter $\eta$. Second, \textsf{DATa} is able to offer state-of-the-art performance even with a much smaller training dataset for real-world applications. Unlike the majority of the state-of-the-art methods, \textsf{DATa} was pre-trained only on one benchmark dataset, named ICDAR 2013, which is one of the smallest, but still exhibits superior performance. Lastly and most significantly, since \textsf{DATa} relies less on the training dataset visually, it is able to perform satisfactorily on transferred domains. Thus, the proposed \textsf{DATa} method is highly applicable to real-world problems more readily and flexibly.}
    
\textcolor{black}{Nevertheless, since hand-crafted lexical features are customizable, they might be less appropriate for certain particular document formats. For more general applications,} potential avenues of future research include the design of a more sophisticated table detection model in transferred domains using deep transfer learning architectures.

\end{document}